\documentclass[sn-mathphys]{sn-jnl}

\usepackage{epsfig}
\usepackage{multirow}
\usepackage{algpseudocode}
\usepackage{amsthm}
\usepackage{amssymb}
\usepackage{amsmath}
\usepackage{bm}
\usepackage{stackengine}
\usepackage{breqn}
\usepackage{float}
\usepackage{mathrsfs}
\usepackage{mathtools}
\usepackage{amsfonts}
\usepackage{graphicx}
\usepackage{algorithm, algpseudocode}
\usepackage{subfigure}
\usepackage{amsfonts}
\usepackage{relsize}
\usepackage{lmodern}
\usepackage{slantsc}
\usepackage[mathscr]{eucal}
\usepackage[final]{pdfpages}
\usepackage{stfloats}
\usepackage{blindtext}
\usepackage{scalerel}
\usepackage{caption}
 \usepackage{multirow}
\usepackage{flushend}
\usepackage{epsfig}
\usepackage{multirow}
\usepackage{amsthm}
\usepackage{amssymb}
\usepackage{amsmath}
\usepackage{bm}
\usepackage{breqn}
\usepackage{float}
\usepackage{mathrsfs}
\usepackage{mathtools}
\usepackage{amsfonts}
\usepackage{graphicx}
\usepackage{algorithm, algpseudocode}
\usepackage{subfigure}
\usepackage{amsfonts}
\usepackage{relsize}
\usepackage{lmodern}
\usepackage{slantsc}
\usepackage[mathscr]{eucal}
\usepackage{dsfont}
\usepackage{bbm}
\usepackage{caption}
\usepackage{tabularx, booktabs}
\usepackage{adjustbox}
\usepackage{colortbl}
\usepackage{lipsum}
\usepackage{scalerel}								\usepackage{siunitx}									\usepackage[font={small}]{caption}

\def\libullet{ \bm{l}_{(i,\bullet)}^{\kern0.2em \scaleto{{[n]}}{5pt}} }

\def\i{^{i}}

\def\h{\bm{h}}

\def\f{\bm{f}}

\def\W{\bm{W}}

\def\l{^{(l)}}
\def\x{\bm{x}}

\def\t{(t)}

\def\i{^{(i)}}

\def\m{\bm{m}}

\def\uuc{_{c}}
\def\ut{^{t}}
\def\nut{^{t-1}}

\def\c{\bm{c}}
\def\U{U}

\def\b{\bm{b}}
\def\uo{_{o}}
\def\W{\bm{W}}
\def\f{\bm{f}}
\def\uf{_{f}}
\def\uuc{_{c}}
\def\a{\bm{a}}
\def\bi{\bm{i}}
\DeclareMathAlphabet{\mathpzc}{OT1}{pzc}{m}{it}
\DeclarePairedDelimiterX{\norm}[1]{\lVert}{\rVert}{#1}

\algnewcommand\Input{\item[\hspace{6pt}\textbf{Input:}]}
\algnewcommand\Output{\item[\hspace{6pt}\textbf{Output:}]}
\algnewcommand\OutputVal{\textbf{output} }
\newcolumntype{Y}{>{\centering\arraybackslash}X}
\DeclareMathSizes{10}{10}{6}{6}

\jyear{2021}%

\theoremstyle{thmstyleone}%
\theoremstyle{thmstyletwo}%

\theoremstyle{thmstylethree}%

\begin{document}

\title[ODW-Assisted Indoor Localization]{Online Dynamic Window (ODW) Assisted Two-stage LSTM Frameworks for Indoor Localization}


\author[1]{\fnm{Mohammadamin} \sur{Atashi}}\email{m\_atashi@encs.concordia.ca}

\author[2]{\fnm{Mohammad} \sur{Salimibeni}}\email{m\_alimib@encs.concordia.ca}

\author*[2]{\fnm{Arash} \sur{Mohammadi}}\email{arash.mohammadi@concordia.ca}

\affil*[1]{\orgdiv{Electrical and Computer Engineering}, \orgname{Concordia University}, \orgaddress{\street{1455 De Maisonneuve Blvd. W.}, \city{Montreal}, \postcode{H3G 1M8}, \state{QC}, \country{Canada}}}

\affil*[2]{\orgdiv{Concordia Institute for Information System Engineering (CIISE)}, \orgname{Concordia University}, \orgaddress{\street{1455 De Maisonneuve Blvd. W.}, \city{Montreal}, \postcode{H3G 1M8}, \state{QC}, \country{Canada}}}


\abstract{\textbf{Purpose:}  Ubiquitous presence of smart connected devices coupled with evolution of Artificial Intelligence (AI) within the field of Internet of Things (IoT) have resulted in emergence of innovative ambience awareness concepts such as smart buildings  and smart cities. In particular, IoT-based indoor localization has gained significant popularity to satisfy the ever increasing requirements of indoor Location-based Services (LBS). In this context,  Inertial Measurement Unit (IMU)-based localization is of particular interest as it provides a scalable solution independent of any proprietary sensors/modules. Existing IMU-based methodologies, however, are mainly developed  based on statistical heading and step length estimation techniques that, typically, suffer from cumulative error issues and have extensive computational time requirements limiting their application for real-time indoor positioning.\\
\textbf{Methods:} To address the aforementioned issues, we propose the Online Dynamic Window (ODW)-assisted two-stage Long Short Term Memory (LSTM)  localization framework. Three ODWs are proposed, where the first model uses a Natural Language Processing (NLP)-inspired Dynamic Window (DW) approach, which significantly reduces the computation time required for implementation of a Real Time Localization System (RTLS). The second framework is developed based on a Signal Processing Dynamic Windowing (SP-DW) approach to further reduce the required processing time of the two-stage LSTM-based model. The third ODW, referred to as the SP-NLP, combines the first two windowing mechanisms to further improve the overall achieved accuracy. Compared to the traditional LSTM-based positioning approaches, which suffer from either high tensor computation requirements or low accuracy, the proposed ODW-assisted models can perform indoor localization in a near-real time fashion with high accuracy.\\
\textbf{Results:} Performances of the proposed ODW-assisted models are evaluated based on a real Pedestrian Dead Reckoning (PDR) dataset. The results illustrate potentials of the proposed ODW-assisted techniques in achieving high classification accuracy with significantly reduced computational time, making them applicable for near real-time implementations.\\}




\maketitle

\section{Introduction} \label{sec:introduction}

Indoor positioning for providing Location-Based Services (LBSs) and/or Proximity Based Services (PBSs) is of significant importance within the context of Internet of Things (IoT). Generally speaking, LBSs/PBSs play a crucial role in emergence of several novel and intriguing IoT concepts such as smart buildings and smart cities~\cite{Bianchi:2019, Zafari:2019}. The user's location, once estimated, can be utilized to provide different LBSs/PBSs including but not limited to targeted advertisements, context aware solutions, automated access, and tenant assistance. Traditional indoor localization technologies, however, require proprietary infrastructure for network design restricting their scalability. Recently, there has been a surge of interest on Local Positioning Systems (LPS) designed using different advanced technologies such as Bluetooth Low Energy (BLE)~\cite{Parastoo:ICASSP, Mohammad:IoT}, Ultra Wide Band (UWB)~\cite{Parastoo:Sensors}, and Inertial Measurement Unit  (IMU) sensors~\cite{Huang:2020,Maheepala:2020,Solano:2020,Luo:2019}. For instance, Received Signal Strength Indicator (RSSI)~\cite{Monfared:2018, Yang:2020, Sadowski:2018} of BLE signals is utilized to estimate the distance between the receiver and the transmitter via the path-loss model. Drastic and random fluctuations of the RSSI values, however, make such systems vulnerable to high level of uncertainty~\cite{Atashi:2019,Malekzadeh:2019}.

To mitigate negative effects of drastic RSSI fluctuations, different localization methods such as fingerprinting~\cite{Shu:2019}, tri-lateration~\cite{Yang:2020} and triangulation~\cite{Monfared:2018} techniques have been developed. Alternatively and capitalizing on widespread  deployment of IMU sensors in smartphones, there has been an ongoing surge of interest on inertial data to estimate the location of smartphones within indoor environments. Relying on portable and locally embedded sensors, deployment of IMU-based systems is considered an efficient alternative approach to localize a user without utilizing external, proprietary, expensive hardware or wearable sensors~\cite{Zou:2017, Atashi:2020}. The  IMU-based positioning is, however, prone to cumulative error but yet is capable of localizing a target without dependence on any external hardware/sensors

\vspace{.025in}
\noindent
\textbf{Literature Review}:
Generally speaking, IMU sensors, widely embedded in common smartphones, consist of $3$-axis accelerometer, $3$-axis magnetometer, and $3$-axis Gyroscope. Among  different localization methods to analyze IMU data for the purpose of indoor positioning, Pedestrian Dead Reckoning (PDR)~\cite{Beni:2020, Poulose:2019} is the most widely-used approach.  In brief, the PDR approach uses heading and step-length estimates at each time to localize an IMU-enabled  smartphone over time. More  specifically, starting from a known position, typically, the PDR technique  recursively estimates the current location of an IMU-enabled device using two major processing tasks, i.e., heading estimation and step detection~\cite{Harle:2013, Kang:2015}. The initial coordinates and the heading of the user are strictly required for accurate PDR. While IMU-based positioning is prone to cumulative error, it can localize a target without dependence on any external hardware/sensor~\cite{Eyobu:2018}.  In PDR approaches, typically, heading is estimated based on the yaw angle, which is obtained using accelerometer, gyroscope and/or the magnetometer data. However, Yaw-based heading estimation is only reliable when the device is in a plenary position. Furthermore, use of magnetometer's heading angle estimation is also unreliable since the magnetic field received by a smartphone can easily be distorted by the effect of surrounding walls and  presence of other magnetic fields in the venue~\cite{Tadayon:2016}.

To address the aforementioned issues, recently, Deep Neural Network (DNN) based solutions~\cite{Hussain:2019,Wang:2019,Feigl:2019,Wagstaff:2018} have gained considerable attention. DNN-based solutions in the context of IMU-based indoor localization are commonly designed based on the Long Short-Term Memory (LSTM) architecture. LSTM-based techniques can also perform  Action Unit (AU) classification (i.e., classification of long step, short step, Left turn, right turn, and/or stop)  and movement classification (such as walking, running, and/or stop). Real-time implementation of existing LSTM-based solutions, however, is impractical due the excessive computational power required for performing  tensor-based Dynamic Windowing (DW). In other words, in conventional LSTM-based indoor localization methods, the AUs used to be splitted either by a fixed window or in an offline fashion. The paper addresses this gap.

\begin{figure}[t!]
\centering
\includegraphics[scale=0.62]{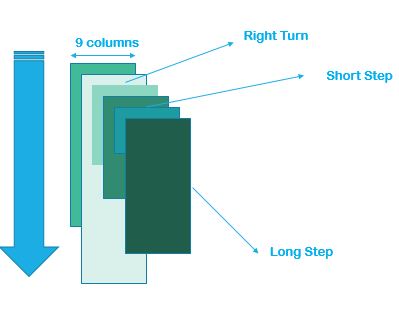}
\vspace{-.25in}
\caption{\normalsize Multivariate AU data. \label{fig:multivariate}}
\end{figure}

\vspace{.025in}
\noindent
\textbf{Contributions:}
In this paper,  we propose a near-real time IMU-based localization framework that measures and validates the bodily acceleration and angular velocity patterns of the subject during indoor movements. The proposed architecture includes two LSTM networks and a moving distance estimator. The two separate LSTMs are trained to learn the distinctive patterns of different movement and AUs. The moving  distance estimator predicts the final position of the subject based on the output of the two LSTMs. Conventional LSTM based indoor positioning solutions~\cite{Wang:2019,Hussain:2019} are not able to estimate the user's location in a real time or at least  in a near-real time fashion. The main objective is to obtain a trade-off between accuracy and latency of the localization system to put one step forward towards having an efficient and respectively accurate indoor localization model. The proposed method consists of the following two phases:
\begin{itemize}
\item\textbf{\textit{Offline Phase}}: First, raw IMU values are collected using the developed iOS SDK (motion collector), which are then smoothed via a moving average filter. The labeled IMU sequential signal is then processed to be fed as the training input to the LSTM classifiers.
\item\textbf{\textit{Online Phase}}: In the online phase, the location of a pedestrian would be estimated based on the performed actions. The inertial IMU data, however, is not splitted into AUs. To resolve this challenge, three innovative Online Dynamic Windowing (ODW) approaches are proposed to receive the sequential IMU data and split it into AUs. The proposed methods rely on advanced Signal Processing (SP) and Natural Language Processing (NLP) techniques.  The proposed SP and NLP inspired ODW mechanisms provide a fair trade off between the accuracy and processing time. An integrated  SP-NLP based ODW is finally proposed to enhance the overall performance of the two-stage LSTM based indoor localization.
\end{itemize}
In addition to the above to phases, to process the real-time sequential IMU data, the LSTM models should be trained on multivariate data stored in the database for each AU. In other words, although the patterns of the data for each AU have high correlation, the length of that AU can vary based on physical parameters of the user's body, frequency of steps taken by the user, and/or the gait cycle information. To address this issue, the proposed LSTMs are trained on multivariate data lengths as shown in Fig.~\ref{fig:multivariate}. Initial results of this research work have appeared in Reference~\cite{Amin:ICASSP}.
\setlength{\textfloatsep}{0pt}

The reminder of the papers is organized as follows: Section~\ref{NSec:2}, provides an overview of the IMU-based indoor localization solutions. In Section~\ref{sec:3}, the proposed  ODW-assisted implementations of a near-RTLS two stage LSTM are represented. Experimental results based on a real dataset are presented in Section~\ref{sec:5} illustrating effectiveness and superiority of the proposed ODW assisted localization. Finally, Section~\ref{sec:6} concludes the chapter.

\section{IMU-based Indoor Localization}\label{NSec:2}
\begin{figure}
\centering
\includegraphics[scale = .5]{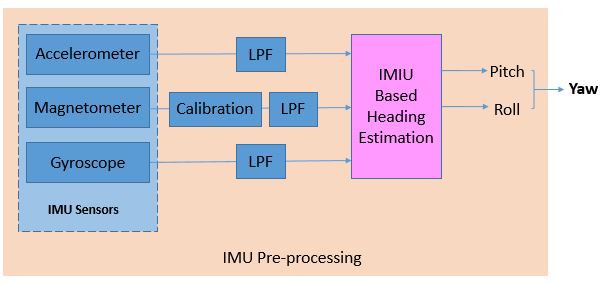}
\caption{\normalsize Heading estimation by IMU.}
\label{fig:IMUdata}
\vspace{.1in}
\end{figure}

Among IMU based indoor localization techniques, PDR is the most widely used approach to iteratively estimate the current location of an IMU-enabled object in an indoor environment. Generally speaking, starting from a known position, successive displacement of the object is estimated via  the following two major steps:
\begin{enumerate}
\item[(i) ]\textit{Step Detection}: The distance travelled by the user holding an IMU enabled device can be represented by her/his number of steps. Therefore, an accurate step detection  algorithm can render better positioning estimation. Although number of steps taken by a user in an indoor environment can be estimated by counting the positive going, zero crossings of a low-pass filtered version of the signal~\cite{Liu:2020,Norrdine:2016,Ruppelt:2015}, the strongest indication of the step specific peak signature is represented on the vertical axis relative to ground~\cite{Kang:2015,Ou:2019}. However, the vertical signal component may be distributed among all three accelerometer axis depending on the present orientation and attitude of the smartphone. To resolve the aforementioned challenge, the axis with highest variation can be selected for step detection evaluation process. Adopting adjacent peak selection, our implemented step detection process is given by
\begin{eqnarray}
\Upsilon^{i} &=& \left\{
\begin{array}{@{}l@{\thinspace}l}
\Upsilon^{i} \ge\tau  \longrightarrow \text{Step} \\
\Upsilon^{i}\le\tau \longrightarrow \text{Local Peak}, \\
\end{array}
\right.
\label{eq:step_detection}
\end{eqnarray}
where the magnitude of consecutive local acceleration peaks ($\Upsilon^{i}$) are subject to a defined threshold ($\tau$), which is an empirically determined constant value. Additionally, to insure a valid global peak (step), the time interval between two consecutive steps should fall between $120$ms to $400$ms. Fig.~\ref{fig:Peak} depicts the estimated steps based on smoothed version of acceleration signal.
\item[(ii) ]\textit{Heading Estimation Unit}: In order to determine the heading of a planar smartphone, first pitch and roll angles are directly calculated based on the accelerometer's readings as follows
\begin{eqnarray}
{P}&=&\arctan \frac{A_y}{\sqrt{{A_x}^{2}+{A_z}^{2}}}
\label{eq:P}
\\
\text{and} \quad  {R} &=&\arctan {\frac {-A_x}{A_z}} .
\label{eq:R}
\end{eqnarray}
Once the pitch and roll angles for each step are calculated, yaw angle can be determined as follows
\begin{eqnarray}
X_h &=& {M_x}\cos{(P)} +{M_y} \sin{(P)} \sin{(R)}+ {M_z} \sin{(P)} \cos{(R)}
\label{eq:Xh}\nonumber\\
\text{and} \quad  Y_h &=& {M_y} \cos{(R)} + {M_z} \sin{(R)},
\label{eq:Yh}
\end{eqnarray}
where
\begin{eqnarray}
P &=& \arctan \{A_y/\sqrt{{A_x}^{2}+{A_z}^{2}}\}\\
R &=&\arctan\{(-A_x)/A_z\} \\
\text{and} \qquad Yaw^{\text{Rad}}&=& \arctan{-(\frac{Y_h}{X_h})}.
\label{eq:yaw}
\end{eqnarray}
In the case that the user holds the device in a plenary position, yaw angle is regarded as the  heading angle of the device. Fig.~\ref{fig:IMUdata} depicts a brief overview of the aforementioned IMU heading estimation approach. Yaw-based representation of heading estimation, in fact, is not practical when the smartphone is swinging in the user's hand or in rests in her/his pocket. To consider the effect of the smartphone's position on the heading estimation algorithm, Principal Component Analysis (PCA) and  PCA-based method coupled with global accelerations (PCA-GA) are employed in conventional PDR-based localization researches. These methods are,  however, still error-prone since magnetometer is vulnerable and easily influenced by interferences caused by external magnetic fields.
\end{enumerate}
\subsection{Machine Learning-based Pattern Recognition} \label{subsec:ML-pattern}
Physical differences of individuals such as height and step length complicates the PDR localization techniques making low pass filtered and smoothed sequential data reported by IMU sensors to be insufficient to distinguish various motion modes. Thus, there is a need for feature extraction from filtered IMU data using a sliding window. The length of the sliding window is typically fixed, which can include several steps. Once the sliding window is applied on the sequential data, it provides an estimated label (motion mode) for each scanned segment of the data. In fact, the periodic features of steps can be modeled in prior steps to predict the occurrence of posterior steps with a higher confidence rate. As shown in previous works, time domain features such as acceleration and velocity's variance and mean values, and frequency domain features such as Short Time Fourier Transform (STFT) are not reliable enough to detect the posterior steps in the path trajectory. The extracted features are Generally speaking, handcrafted features fail to sufficiently represent the input data, as such it is, typically, challenging to identify representative handcrafted features for classification tasks at hand. The most recent articles, therefore, have gone beyond the conventional feature extraction solutions to extract deep features from the sequential inertial signal.

\subsubsection{Long Short Term Memory (LSTM) Architecture}\label{2StageLSTM}
As stated previously, analyzing the gait information in IMU data, i.e., step and heading based indoor localization techniques, have strict limitations making them non reliable positioning systems. To address the aforementioned challenges from a practical  perspective, deep models have been emerged. Typically, DNN methods render a structure in which end-to-end learning, automated feature extraction and classification, are performed jointly instead of using handcrafted features. The essence of IMU-based localization is developed on the bases of recursive plausible location estimation via prior step coordinates, i.e., smoothed $3$-axis accelerometer and gyroscope data form sequential time series. Recurrent Neural Networks (RNN) with linear chin structure has been deployed to analyze time sequence data in various domains, e.g., automation, NLP, speech recognition, image captioning and handwriting recognition. Despite the numerous accomplishments made by RNN networks,  the limited capacity of contextual information and inability to back propagate in time has been reported as the downfalls of such networks. Since RNN consists of iterative processing of the data segments, such networks and its variants are prone to vanishing and exploding gradient problems. LSTM is a type of artificial Recurrent Neural Network, deployed to address the aforementioned problems~\cite{Hussain:2019,Wang:2019,Feigl:2019,Wagstaff:2018}. A typical LSTM unit is consist of a memory cell, an input gate, an output gate, and a forget gate. The cell in LSTM is designed to process sequential segments of the data and maintains its hidden state through the course of learning. The implementation of cell assists LSTM network to overcome the challenges of traditional RNN during learning process.

More specifically, the LSTM takes as input a single time window and learns to model the underlying sequence based on its corresponding label. Typically, LSTM's structure is many-to-one  and the input time-window is of fixed length. The output of the LSTM cell is denoted by $\h\ut$$\in \mathbb{R}^{\m \times 1}$, where $\m$ is the number of nodes. Similarly,  the LSTM's  cell state is denoted by $\c\ut$$\in \mathbb{R}^{\m \times 1}$. At each time step $t$, LSTM receives the sensor data $\x\ut$ together with the output $\h\nut$ and the hidden state $\c\nut$ from the previous time step. The LSTM layer, at each time step $\t$, is implemented based on the following formulation
\begin{eqnarray}\label{eq:LSTM1}
\h\ut &=& \o\ut \circ \tan(\c\ut) ,\label{eq:LSTM6}\\
\text{where} \quad \o\ut &=& \sigma (\W\uo \x\ut + \U\uo \h\nut + \b\uo) ,\label{eq:LSTM2}\\
\text{and} \quad \c\ut &=& \f\ut \circ \c\nut + \bi\ut \circ \a\ut  , \label{eq:LSTM5}\\
\text{with} \quad \f\ut &=& \sigma (\W\uf \x\ut + \U\uf \h\nut + \b\uf), \label{eq:LSTM3}\\
 \bi\ut &=& \sigma (\W\i \x\ut + \U\i \h\nut + \b\i) ,\\
\text{and} \quad \a\ut &=& \tan(\W\uuc \x\ut + \U\uuc \h\nut + \b\uuc) , \label{eq:LSTM4}\\
\end{eqnarray}
where $\o\ut$$\in \mathbb{R}^{\m \times 1}$ is the output gate controlling the information to be forwarded in time; $\bi\ut$$\in \mathbb{R}^{\m \times 1}$ is the input gate that selects the content to be sent to the memory cell, and;  $\f\ut$$\in \mathbb{R}^{\m \times 1}$ is the forget gate that controls the update process of the memory cell. Furthermore, $\W\i$, $\W\uo$, $\W\uf$, $\W\uuc$ $\in \mathbb{R}^{\m \times \l}$, and $\U\i$, $\U\uo$, $\U\uf$, $\U\uuc$ $\in \mathbb{R}^{\m \times \m}$ are weight matrices; Terms $\b\i$, $\b\uo$, $\b\uf$, $\b\uuc$$\in \mathbb{R}^{\m \times 1}$ are bias vectors; $\sigma(\cdot)$ denotes the sigmoid activation function; Term ``$\circ$" represents the Hadamard product (i.e., element-wise multiplication of two vectors), and; $\tan(\cdot)$ represents element-wise hyperbolic tangent activation function.

\section{Proposed ODW assisted Two Stage LSTM Architecture}\label{sec:3}
\begin{figure} [t!]
\centering \includegraphics [width=0.75\textwidth] {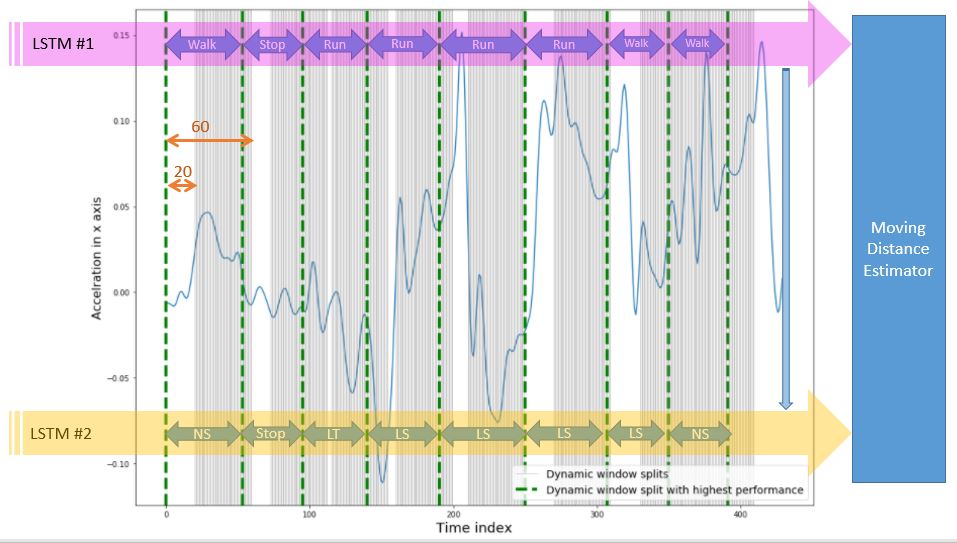}
\caption{\normalsize Conventional two stage LSTM.} \label{fig:conventional_LSTM}
\vspace{.3in}
\end{figure}
Recently there has been attempts to enhance the accuracy of indoor positioning using RNN, LSTM and its variants~\cite{Hussain:2019}. Great number of such systems utilize IMU of smartphone to measure bodily acceleration and angular velocity associated with different AUs. As depicted in Fig.~\ref{fig:conventional_LSTM}, such systems include two LSTM classifiers and a moving distance estimator. While the first LSTM classifies the user's movement state (i.e., stop walking, running), the second one is designed to recognizing the AUs performed by the user (i.e., left and right turn, short, normal and long step, abnormal activity). However, the accelerometer and gyroscope readings representing bodily acceleration and angular movements of the subject in Cartesian coordinates, are prone to drastic fluctuations in a sample course of time, most of which can be smoothed using a moving average filter.

\begin{figure}[t!]
\centering
\includegraphics[scale=0.40]{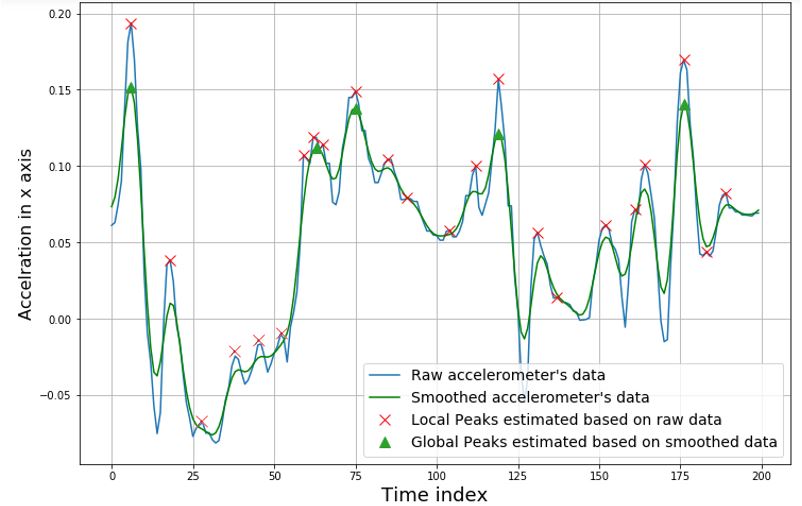}
\vspace{-.1in}
\caption{\normalsize Raw and smoothed inertial reports and illustration of a peak detection algorithm.  \label{fig:Peak}}
\end{figure}

\vspace{.025in}
\noindent
\textbf{Moving Average (Smoothing) Filter:} In order to mitigate, and if applicable remove the level of drastic fluctuations in raw inertial reports of $3$-axis accelerometer, magnetometer and gyroscope, the data is smoothed using a moving average filter given by
\begin{equation}
{x}_{f}(n)_{\alpha}=\frac{1}{N} \sum_{k=0}^{N-1} x_{\alpha} ({n}+{k});
\\
\alpha=\{a_{x,y,z}, g_{x,y,z}\},
\label{eq:smoothing_filter}
\end{equation}
where $\alpha$ represents the bodily acceleration $a_{x,y,z}$ and angular movements $g_{x,y,z}$ of the subjects in the  $X$, $Y$, and $Z$  axes, respectively. Term $N$ is the number of indexes in a sample inertial signal. Fig.~\ref{fig:Peak} illustrates the raw and smoothed versions of a sample accelerometer's data. The smoothed inertial data can be used as the input to  the signal processing (i.e., step and zero cross point detection) and the pattern recognition (i.e., RNN, LSTM-based AU classification) modules.

After applying the moving average filter, the smoothed IMU data is  labeled empirically in an offline phase. The labeled $6$ axis inertial data (consisting $3$ axis accelerometer and $3$ axis gyroscope data) is then fed to the LSTM network as the training data. It is worthy to mention that once the smoothed inertial data is derived, the training and test segments of data are splitted in an offline phase using a dynamic window given by
\begin{eqnarray}
A&=&\begin{pmatrix}
&S_1(x_f^a(n+L)&\\
&S_2(x_f^a(n+L)&\\
&\vdots& \\
&S_{i}(x_f^a(n+L)&
\end{pmatrix}
(20\leq{L}\leq60),
\label{eq:conventional_DW}
 \end{eqnarray}
where {$S_i(x_f^a(n+L))$} represents the $i^{th}$ segment of the filtered inertial signal. The segmented inertial sensor data reported by IMU sensors carry sufficient information about the bodily acceleration and angular velocity patterns of different AUs and movements performed by the user during the path trajectory. Different data segments along with the empirically derived labels will be sequentially fed to LSTM networks, which classify segments into different movement states (i.e., running, walking, stopping) and different AUs (i.e., left and right turn, short, normal and long step, abnormal activity). The challenge preventing the algorithm to be identified as a real-time localization is the dynamic windowing mechanism, which excessively splits the sequential data into numerous data segments, which should be fed to LSTM network in an offline phase. The algorithm would then choose the most viable segment based on the best recognition performance.

The implementation principle of moving distance estimator is similar to PDR, although in the proposed two stage LSTM, the current location of the user is updated based on each recognized activity and its related AU rather than heading and statistical stride length estimation. The moving distance estimator is given by $P_k = P_k-1 +A{\amalg_k}$, where $k$ shows the current time index of the position ($P$) and its corresponding  recognized action unit (A{$\amalg_k$}). The step length of the participants using the mode length value of step type recognized by the LSTM (e.g., SS, NS, or LS) can be determined as follows
\begin{eqnarray}
A{\amalg_k} &=& {\Psi_k},\\
\text{and} \quad {\Psi_k} &=& \Pi({\gamma_k},{\lambda_k},{\delta_k}),
\end{eqnarray}
where {$\Psi_k$} is the assumed step length, and $\Pi$ is the function mapping each subject ({$\gamma_k$}) with the mode of step length ($\gamma_k$) of the particular step length type ({$\lambda_k$}) according to type of activity ($\delta_k$) performed in the current time index ($k$). Once the most viable segment based on the best performance of the recognition model was chosen, then the next segment is fed to LSTM network. The implementation of excessive tensor computation is not efficient in terms of time, memory and computation power. The conventional dynamic window for two stage LSTM indoor positioning does not satisfy the RTLS requirements. Although the algorithm yields high accuracy of AU classification, the dynamic window slides through the test data in an offline phase requiring excessive computation for each AU recognition. Even if implemented in semi real-time running manner, there is lack of processing power to deal with high requirement s of tensor computation and AU recognition. The aforementioned challenges are targeted via the following proposed three different (SP, NLP and SP-NLP based) ODWs.

\subsection{NLP Inspired Dynamic Window} \label{subsec:NLPDW}
\begin{figure}[t!]
\centering
\includegraphics [scale = 0.45] {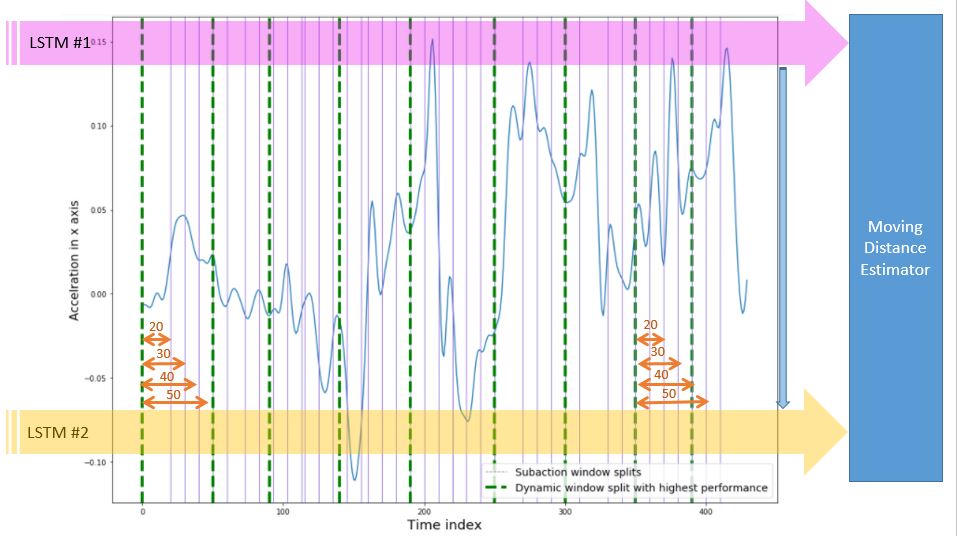}
\caption{\normalsize NLP inspired online dynamic window.} \label{NLP_inspired}
\vspace{.3in}
\end{figure}
Inspired by ubiquitous implementation of NLP techniques in various domains, a solution for near real-time implementation of two stage LSTM is proposed. Considering  similarities between sequential IMU and text data, we can model the positioning problem in terms of NLP, where the algorithms can be trained based on multivariate sentences as input data. Once trained, these frameworks can perform classification on multivariate sequential data in the test phase. In fact, in order to process the real time sequential data collected from iOS Software Development Kit (SDK), the LSTM models should be trained on multivariate data stored in the database (Fig.~\ref{fig:multivariate}) for each AU in the database. In other words, although the patterns of the data for each AU are highly correlated, the length of that AU can vary based on physical parameters of the user's body, frequency of steps taken by the user and the gait cycle information of the pedestrian. The ``multivariate'' term in sequential data refers to the non-uniformly splitted subsets of time series data in this research work. In fact, a typical statement consist of several sentences, each of which includes different number of words. In order for the algorithms to be trained on the multivariate data a practical technique is deployed to turn meaningful pieces of data (such as words) into random string of characters or numbers called ``tokens'' such that no meaningful value is breached. Therefore, the problem of training and testing multivariate data is resolved by the tokenization technique. Tokenization is considered as a key and mandatory aspect of working with text data in NLP applications.

Considering the aforementioned solution to resolve multivariate training of the NLP models, we model IMU sequential readings to the statements in NLP. Additionally, deploying the tokenization concept for AU classification, we consider tokens as small subsets of an action performed by the user. Similar to NLP classification methods, where the tokenized words as subsets of a statement form a sequential meaningful statement, in this application uni-variate subsets of AUs form a meaningful action performed by the user. Implementation of such method prevents excessive tensor calculations and reinforces the performance of near real-time indoor tracking method. In other words, the proposed NLP-based dynamic window would split the test set into considerably fewer number of tensors. Consequently, the algorithm can assess the performance of the LSTM network with different input lengths (multiplies of token length) and recognizes the user's AU. As is shown in Fig.~\ref{NLP_inspired}, the NLP inspired DW can split the sequential test data with fewer number of segments, which reduces the computation requirement of the proposed framework.

\subsection{Signal Processing Dynamic Window} \label{subsec:SPDW}
\begin{figure} [t!]
\centering \includegraphics [scale = 0.55] {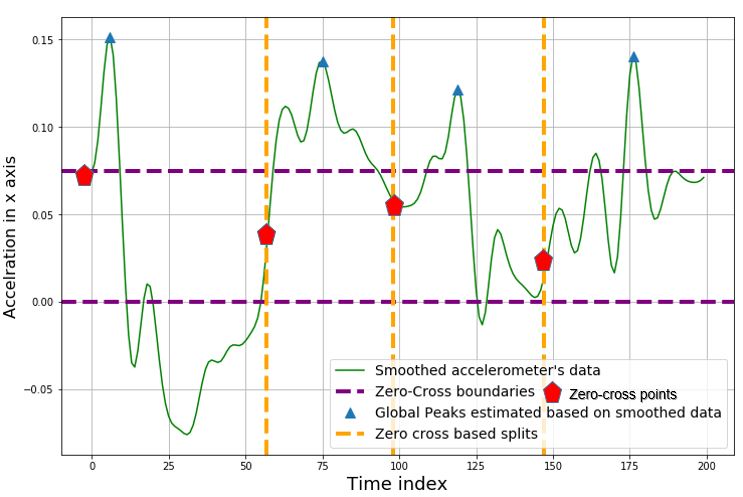}
\caption{\normalsize Signal Processing based online dynamic window.} \label{Fig:SP}
\vspace{.3in}
\end{figure}

As another solution to establish a real-time indoor localization, we attempt to analyse the signal pattern using an accurate step detection method to split the test data based on the knowledge of the step indexes.
Once the step indexes are identified, the SP-DW searches for zero crossing points that identifies the starting and ending points of each performed AU.  For this purpose, a rule based SP dynamic window is proposed to detect the starting and end point of each AU. Each zero crossing point should abide all of the following rules to be picked as starting/end index of an AU. First of all, each zero-crossing point should be between two consecutive peak points. Moreover, each zero point should be in zero-crossing vicinity. In the normal walking mode, each zero-crossing point should be almost in the middle of two consecutive peaks. Additionally, there should not be two zero-crossing points between two consecutive peaks. The pedestrian steps are represented through the distinct peak patterns, as shown in Fig.~\ref{Fig:SP} where the number of peaks indicate the total number of steps. Although SP-dynamic window is not able to accurately recognize the starting and end of AUs, the required computational delay in this method is less  than that of the conventional and NLP-based dynamic windows since its performance is not dependent on any tensor assessment. In other words, SP-dynamic window attempts to split the test data and recognizes the starting and end of each AU without trying to identify the action. Once the test data is splitted, the AUs will be transformed to LSTM classifier to detect the corresponding label of each AU based on the signal pattern and the deep extracted features of the model.

\subsection{SP-NLP Fusion Dynamic Window}\label{fusionDW}
\begin{figure}[t!]
\centering \includegraphics [scale = 0.55] {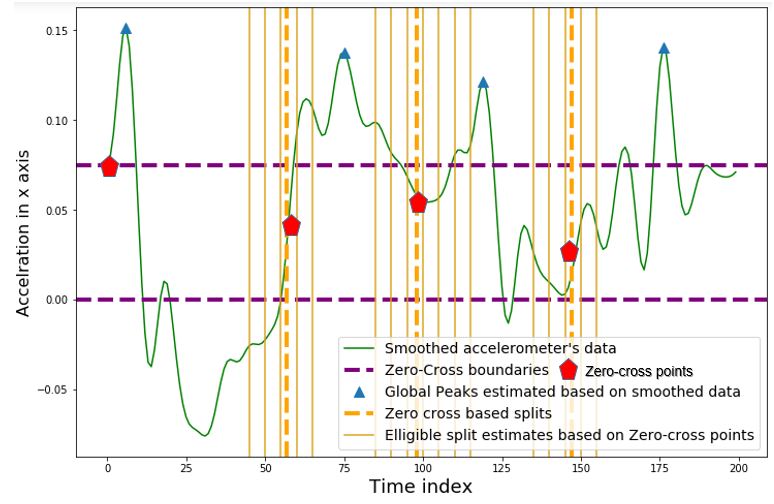}
\caption{SP-NLP based online dynamic window.} \label{SP_NLP}
\end{figure}

The proposed SP and NLP methods, each benefit from particular and different advantages for real-time implementations. The methodology of NLP-based dynamic window is based on fewer number of tensor calculations and the assessment of LSTM model on multivariate test data, while the SP-based dynamic windowing approach provides a much faster test splitting technique. In real time scenarios, the SP-based dynamic window renders higher localization speed with lower accuracy while the NLP approach takes advantage of multiple tensor assessments prior to AU classification. Such differences in performance of real-time scenarios led to deployment of a fusion model referred to as the SP-NLP fusion dynamic window. The goal of such fusion model is to establish a trade off between accuracy and speed of the RTLS. The  proposed SP-NLP fusion dynamic window consists the following two phases:
\begin{enumerate}
\item[(i)] \textbf{\textit{SP-DW}}: First, the SP framework would detect the peaks and eligible step indexes in the sequential preprocessed signal reported by IMU sensors. Based on the peaks and rule-based zero crossing vicinity, the zero-crossing points in the sequential data would be detected. By determining the zero-crossing indexes, the sequential data can be splitted into segments to be provided as inputs to LSTM classifiers. If the accuracy of the LSTM classification exceeds a predefined threshold, the segment would be picked as an AU. Otherwise NLP-DW would be used to segment the AU in a more accurate fashion. In other words, the algorithm evaluates the accuracy of SP-DW segments ($\Upsilon$), with regards to a predefined threshold $\tau_1$. Then the framework would either implement the localization based on SP-DWs or pass the sequential signal to NLP-DW for re-segmentation.
\item[(ii)] \textbf{\textit{NLP-DW}}: If the segments splitted by the SP-DW does not exceed the threshold, an NLP based model would receive the knowledge of the zero crossing indexes in the sequential data and attempts to find the nearest $k$ neighbor tokens of that step. Consequently, $k+1$ tensors would be fed to the  LSTM to recognize the AUs and their corresponding labels. Finally, the segment with highest performance metric (i.e., accuracy) represents the AU. The overall decision algorithm for the proposed algorithm is given by
\end{enumerate}
\[
  \Upsilon = \left\{
     \begin{array}{@{}@{\thinspace}l}
        \Upsilon \ge\tau_1 \longrightarrow \text{Localization}  \\
        \\
        \\
      \Upsilon\le\tau_1  \longrightarrow  \text{NLP-DW} \left\{
      \begin{array}{@{}l@{\thinspace}l}
     \eta\ge\tau_2 \longrightarrow \text{Localization}\\
     \eta\le\tau_2  \longrightarrow \text{No valid AU}
     \end{array}
  \right.

     \end{array}
  \right.
\]
As expected, the accuracy of the SP-NLP dynamic window is higher than the two other  methods (i.e., SP and NLP based DWs) since it simultaneously benefits from not only the zero-crossing indexes but also the tensor calculation. Moreover, in the SP-NLP dynamic window, as the tensors are chosen based on zero-crossing knowledge, the probability of correct AU recognition would be higher. Furthermore, compared to the NLP-based dynamic window, the SP-NLP model would recognize the AUs much faster than proposed and conventional models. Fig.~\ref{SP_NLP} represents the segments of the SP-NLP-based online dynamic windowing approach.

\section{Simulation Results} \label{sec:5}
\begin{figure} [t!]
\centering \includegraphics [scale = 0.51] {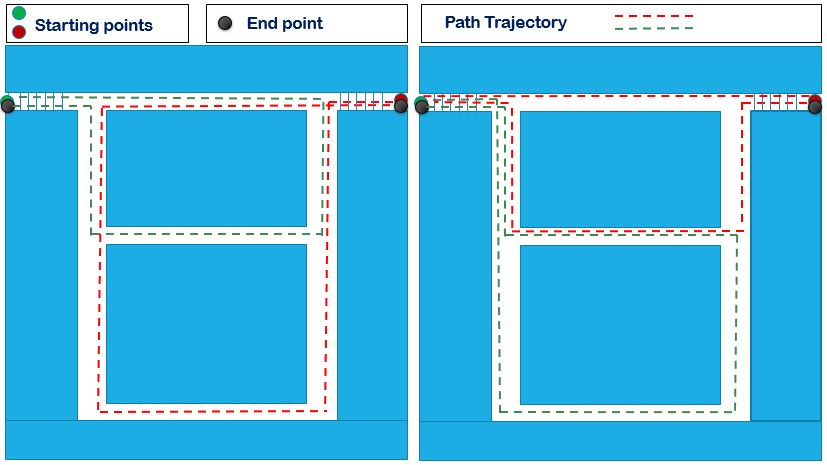}
\caption{\normalsize Path Trajectories designed for ODW assisted two stage LSTM approach.} \label{fig:path}
\end{figure}
To evaluate the real-time implementation of NLP-inspired and SP-based dynamic windows, an experiment protocol is designed prior to data gathering phase. The data used in this research work is consist of data set collected by Ghulam Hussain \textit{et al.} in 2019~\cite{Hussain:2019} and the newly collected data using our developed iOS application and iPhone 11 pro IMU sensors for more comprehensive investigation. In the newly designed data gathering setup, total number of 80 inertial data sets were collected by two different users in 4 distinctive path trajectories illustrated in Fig.~\ref{fig:path}. Each user can follow the pre-defined path trajectories by their own choice. Based on the predefined instructions, in half of the test data sets (40 tests) the user holds the smartphones by their right hand and the rest of the data was collected while the user holds the smartphone by left hand. To be aligned with the conventional data set the sampling frequency was fixed (50Hz). The inertial raw data collected (Comma-Separated Values (CSV)) files in the smartphone is propagated to the back-end server using a smartphone SDK. As in real time scenarios, the movement statuses are not confined to stop walking and running, the newly designed path trajectories consist two more movement statements of upstairs and downstairs to further enhance the indoor localization technique.

\begin{figure*} [t!]
\centering \includegraphics [scale = 0.54] {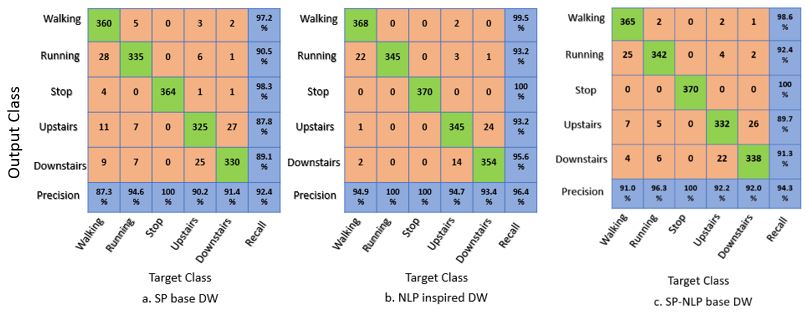}
\caption{\normalsize Confusion matrix comparison of different DWs representing the LSTM performance for  Movement status recognition (test accuracy)} \label{fig:moving_states}
\end{figure*}
\begin{figure*} [t!]
 \includegraphics [scale = 0.44] {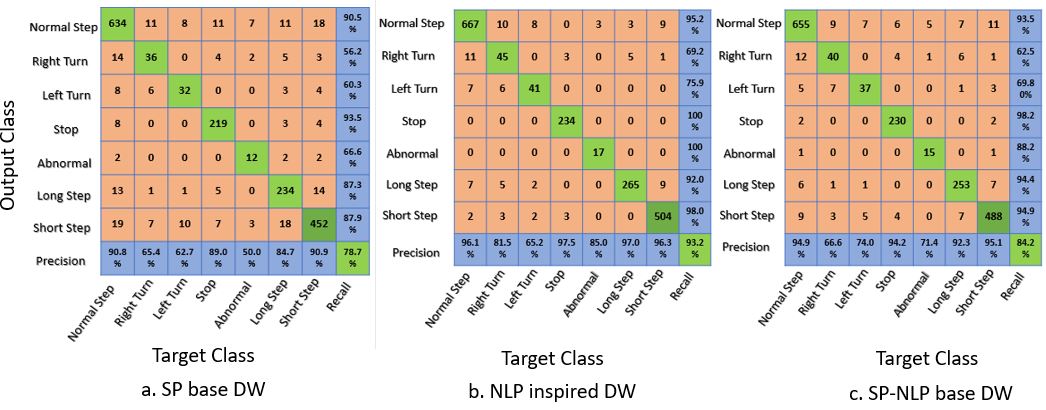}
\caption{Confusion matrix comparison of different DWs representing the LSTM performance  for AU recognition (test accuracy)} \label{fig:ConfusionV1}
\end{figure*}
The corresponding AU label to each time segment is reported by actively monitoring pedestrian trajectory using cameras installed in the venue.
In this experiment, $7$ AUs (Long Step (LS), Normal Step (NS), Short Step (SS), Left Turn, Right Turn, Abnormal, Stop) and $5$ moving states (Walking, Running, Stop, Down Stairs, Upstairs) are considered for LSTM classifications. The LSTMs were trained over the data on conventional research work as well as the newly collected data after initialization of hyper parameters. Once the LSTM networks are trained, the implementation of different proposed dynamic windows can be evaluated on the test set. Despite the conventional LSTM based positioning systems, the test set in this chapter is not splitted into AU in a passively in an offline manner. In contrast to contemporary LSTM approaches, in this implementation, the algorithm receives the test set of data simultaneously as the user walks through path trajectory in the venue. Assisted by proposed dynamic windows, the methods attempt to split the test data stream into eligible subsets representing AUs. Table
\ref{tab:SP-NLP-comp} provides the run time and average accuracy comparison of ODW assisted two stage LSTM based indoor localization. As expected, the required time to process an AU in proposed methods (SP-DW and NLP-DW) are considerably less than the conventional model. More importantly, the accuracy of AU and moving state classification (LSTM 1 and LSTM 2) remained respectively high. NLP model is more accurate in positioning since the distance in such model is obtained from high tensor computations whereas in NLP inspired DW model, an average error of \SI{0,934}{\metre} in an indoor area of \SI{131,5}{\metre\squared} was measured. The confusion matrix comparison of conventional, SP based and NLP inspired DWs are reported in Figs.~\ref{fig:moving_states} and \ref{fig:ConfusionV1}.
\begin{table*}[t!]
\centering
\caption{\normalsize Run time and average accuracy comparison of ODW assisted two stage LSTM based indoor localization.}
\vspace{.1in}
\label{tab:SP-NLP-comp}
\centering \includegraphics [scale = 0.75] {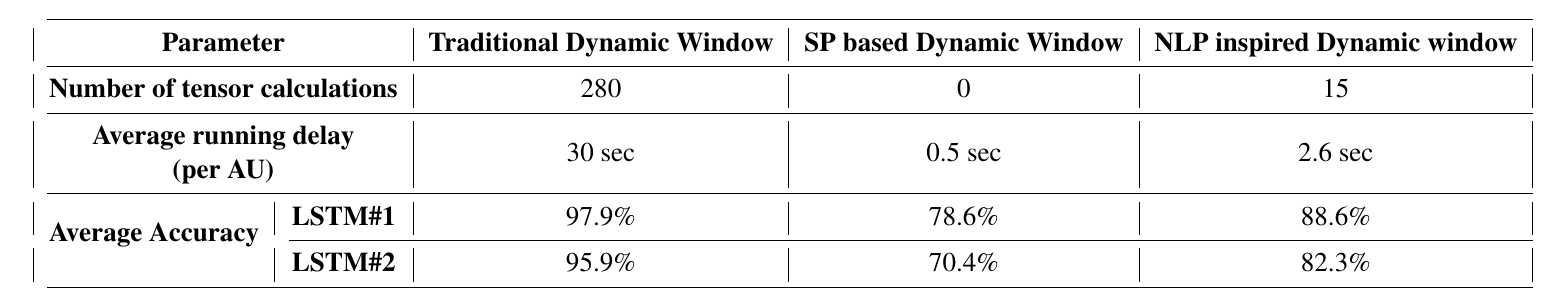}
\end{table*}
\begin{table*}[t!]
\centering
\caption{\normalsize Comparison of different localization approaches.}
\vspace{.1in}
\label{tab:SP-NLP-comp}
\centering \includegraphics [scale = 0.75] {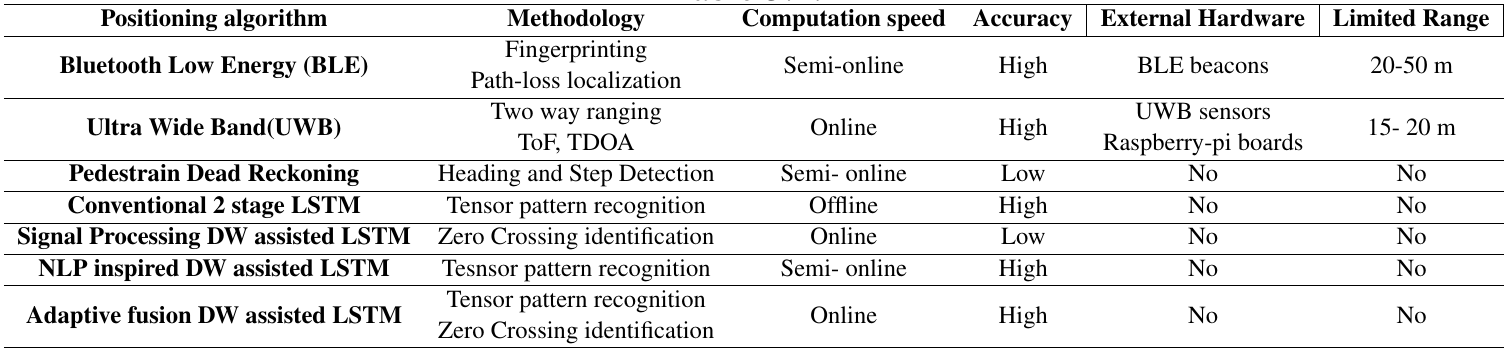}
\end{table*}

\section{Conclusion} \label{sec:6}
The paper proposed a novel Online Dynamic Window (ODW)-assisted two-stage LSTM framework for near real-time localization based on distinctive IMU data patterns. First, NLP inspired DW is introduced which could significantly reduce the computation time required for indoor positioning. Second, to analyze the IMU sequential signal, SP-DW was implemented, which could further decrease the processing time. Finally to establish a trade of between the accuracy and the running time, SP-NLP-based ODW is proposed, which combines the first twos categories. The proposed framework consists of two LSTM classifiers and a moving distance estimator. The first LSTM classifies the user's movement state (i.e., stop, walking, and running states) while the second LSTM model is designed to recognize the AUs performed by the user (i.e., left and right turn, short, normal and long step, and abnormal activity). The moving distance estimator updates the current location of the user based on each recognized activity and its associated AU. The proposed framework consists of the following two phases: (i) \textit{Offline Phase}: The IMU values for each AU and movement state is collected, smoothed via moving average filter, and labeled empirically using the video cameras installed in the venue. (ii) \textit{Online Phase}: This phase consists of pattern matching and post-processing steps. In the online phase, the real-time sequential IMU data is measured, then a moving average filter is applied for two reasons: (i) To smooth fluctuations and sudden drifts in the IMU signals, and; (ii) To prepare the data for step, peak, and zero cross detection algorithms. The pre-processed sequential data would then be splitted using the proposed ODW methodologies (i.e., inspired by NLP and SP techniques). The proposed IMU-based indoor localization approach can be further improved by extending the two-stage LSTM framework to consider other embedded sensors of the smart-phone such as lightening and/or barometer sensors.

\vspace{6pt}

\backmatter

%
%

\bmhead{Acknowledgments}
This work was partially supported by the Natural Sciences and Engineering Research Council (NSERC) of Canada through the NSERC Discovery Grant RGPIN-2016-04988.

%
%


\end{document}